\documentclass[10pt,twocolumn,letterpaper]{article}

\usepackage{iccv}
\usepackage{times}
\usepackage{epsfig}
\usepackage{graphicx}
\usepackage{amsmath}
\usepackage{amssymb}

\usepackage{subcaption}
\usepackage{multirow}
\usepackage{booktabs}
\usepackage[table]{xcolor}
\usepackage[breaklinks=true,bookmarks=false]{hyperref}
\usepackage{cleveref}
\iccvfinalcopy % *** Uncomment this line for the final submission

\def\iccvPaperID{} % *** Enter the ICCV Paper ID here

\ificcvfinal\fi

\begin{document}

%%%%%%%%% TITLE
\title{Evolving from Unknown to Known: Retentive Angular Representation Learning for Incremental Open Set Recognition}

\author{Runqing Yang\\
{\tt\small nwpuyrq@mail.nwpu.edu.cn}
\and
Yimin Fu\textsuperscript{*}\\
{\tt\small fuyimin@hkbu.edu.hk}
\and
Changyuan Wu\\
{\tt\small wuchangyuan@mail.nwpu.edu.cn}
\and
Zhunga Liu\\
{\tt\small liuzhunga@nwpu.edu.cn}
}

\maketitle
\ificcvfinal\thispagestyle{empty}\fi

%%%%%%%%% ABSTRACT
\begin{abstract}

Existing open set recognition (OSR) methods are typically designed for static scenarios, where models aim to classify known classes and identify unknown ones within fixed scopes.
This deviates from the expectation that the model should incrementally identify newly emerging unknown classes from continuous data streams and acquire corresponding knowledge.
In such evolving scenarios, the discriminability of OSR decision boundaries is hard to maintain due to restricted access to former training data, causing severe inter-class confusion.
To solve this problem, we propose retentive angular representation learning (RARL) for incremental open set recognition (IOSR). 
In RARL, unknown representations are encouraged to align around inactive prototypes within an angular space constructed under the equiangular tight frame, thereby mitigating excessive representation drift during knowledge updates. 
Specifically, we adopt a virtual-intrinsic interactive (VII) training strategy, which compacts known representations by enforcing clear inter-class margins through boundary-proximal virtual classes.
Furthermore, a stratified rectification strategy is designed to refine decision boundaries, mitigating representation bias and feature space distortion caused by imbalances between old/new and positive/negative class samples.
We conduct thorough evaluations on CIFAR100 and TinyImageNet datasets and establish a new benchmark for IOSR.
Experimental results across various task setups demonstrate that the proposed method achieves state-of-the-art performance.
\end{abstract}

%%%%%%%%% BODY TEXT
\section{Introduction}

\label{sec:intro}
Open set recognition (OSR)~\cite{scheirer2012toward} is recognized as an effective approach to enhancing model reliability, driving advances across various tasks, including face recognition~\cite{gunther2017toward}, multimodal perception~\cite{fu2024logit}, and fault diagnosis~\cite{yu2021deep}.
However, most current studies are designed for static scenarios, remaining limited to identifying known classes within a fixed scope while identifying unknowns. In contrast, a more desirable expectation for OSR methods is to leverage the identified unknown classes for knowledge update, thereby continuously expanding the recognizable scopes in evolving scenarios.
We term the aforementioned problem incremental open set recognition (IOSR), and requires the model to possess the following two abilities:
(1) accurately classifying known classes while effectively rejecting unknown ones, and (2) continuously integrating new classes into the recognizable scope while preserving previously learned knowledge.

\begin{figure}[]
    \centering
    \begin{subfigure}{\linewidth} % 指定子图的宽度
        \centering
        \includegraphics[width=\linewidth]{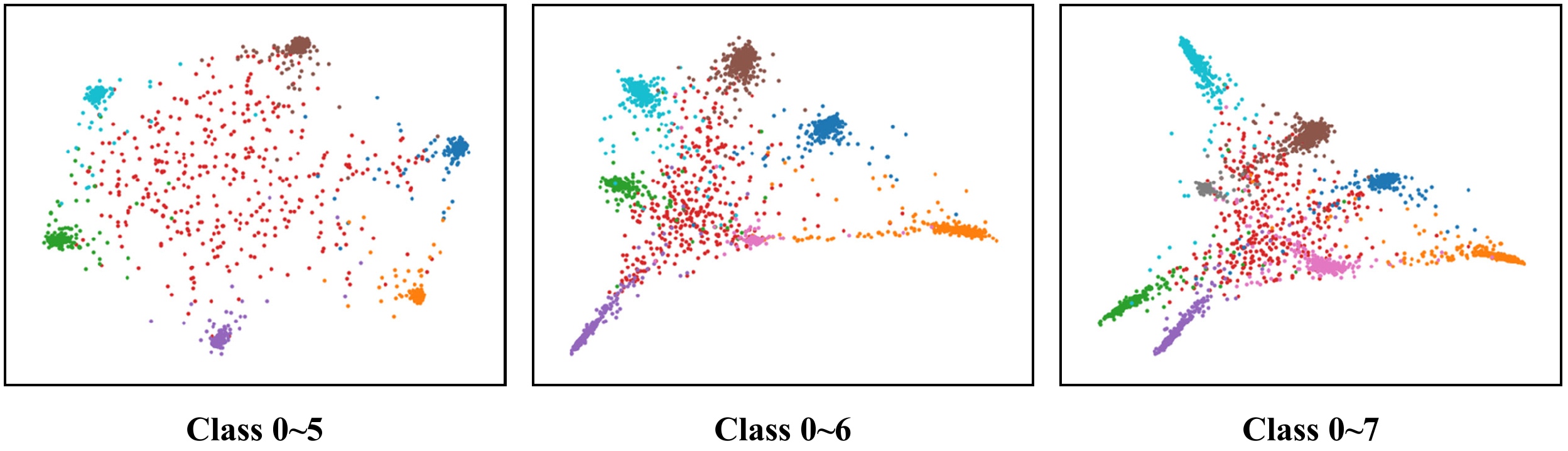}
        \caption{Training with limited old data}
        \label{Figure1a}
    \end{subfigure}
    \\
    \begin{subfigure}{\linewidth}
        \centering
        \includegraphics[width=\linewidth]{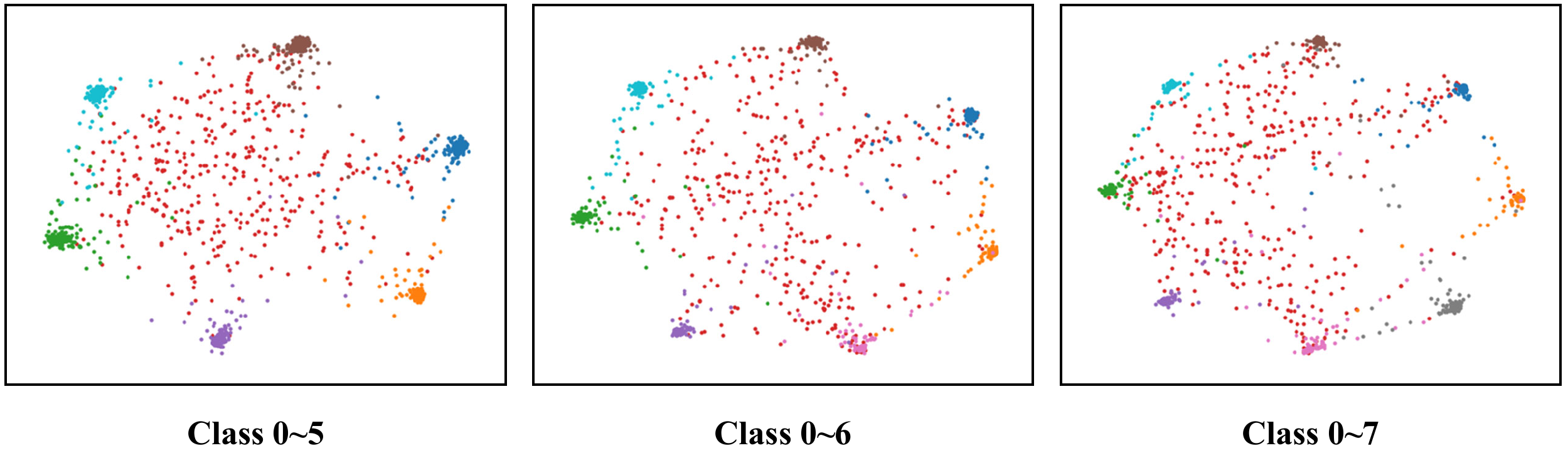}
        \caption{Joint training using all data}
        \label{Figure1b}
    \end{subfigure}
    \caption{Feature distribution on the MNIST dataset under evolving scenarios, where red dots represent unknown classes.}
    \label{Figure1}
\end{figure}

\begin{figure*}[t]
    \centering
    \begin{subfigure}{0.24\textwidth} % 指定子图的宽度
        \includegraphics[width=\linewidth]{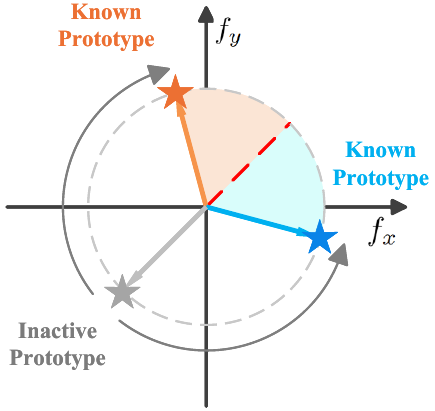}
        \caption{}
        \label{Figure2a}
    \end{subfigure}
    \hfill
    \begin{subfigure}{0.24\textwidth}
        \includegraphics[width=\linewidth]{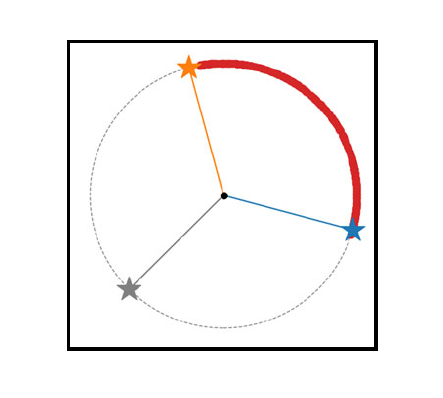}
        \caption{}
        \label{Figure2b}
    \end{subfigure}
    \hfill
    \begin{subfigure}{0.24\textwidth}
        \includegraphics[width=\linewidth]{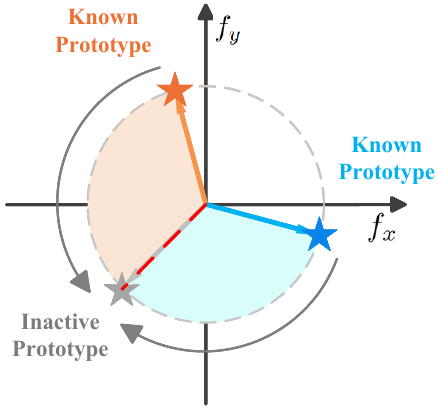}
        \caption{}
        \label{Figure2c}
    \end{subfigure}
    \hfill
    \begin{subfigure}{0.24\textwidth}
        \includegraphics[width=\linewidth]{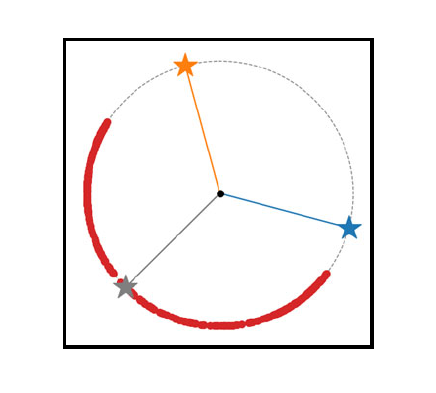}
        \caption{}
        \label{Figure2d}
    \end{subfigure}
    \caption{Feature space for binary classification: The vertices of the maximal ETF in the 2D plane are used as classification prototypes.}
    \label{Figure2}
\end{figure*}

%Existing OSR methods can be mainly categorized into discriminative~\cite{yang2018robust,miller2021class,liu2022orientational,li2024all} and generative~\cite{chen2021adversarial,huang2022class,jang2023teacher} methods, both of which have been well validated in distinguishing between known and unknown classes in static settings.
%Discriminative methods focus on optimizing the latent representations of known classes, whereas generative methods aim to estimate the potential distribution of unknown classes.
When continuous learning from streaming data, storing the training data of all seen classes and retraining the model at each step is impractical due to computation and memory costs.
In contrast, as shown in \cref{Figure1}, the representations of learned known classes tend to drift when the model is updated with limited access to former training data. 
Consequently, the discriminability of OSR decision boundaries is hard to maintain, causing severe confusion both among known classes and between known and unknown classes.
Therefore, the effective mitigation of such representation drift is crucial when evolving open set classes from unknown to known, which is consistent with the objective of class incremental learning (CIL)~\cite{zhou2024class}.

%class incremental learning (CIL)~\cite{zhou2024class} seeks the plasticity-stability balance through three primary strategies: data replay~\cite{rebuffi2017icarl,zhao2021memory,sun2023exemplar}, parameter regularization~\cite{kirkpatrick2017overcoming,yang2019adaptive,hu2021distilling}, and model rectification~\cite{hou2019learning, zhao2020maintaining, he2024dyson}, to prevent catastrophic forgetting of previous knowledge.
Among existing CIL methods, an effective strategy is to pre-fix class prototypes in an equiangular tight frame (ETF)~\cite{papyan2020prevalence, yang2023neural, he2024dyson}.
Although the representations of previously learned classes can be anchored by fixed prototypes, the conventional incremental learning process, which operates under the closed-set assumption, overlooks the identification of unknown classes.
Considering a binary-class 2D feature space, where all vertices within the maximal ETF are set as class prototypes (two known and one inactive).
As demonstrated in \cref{Figure2a}, existing ETF-based CIL methods typically incorporate inactive prototypes into the optimization process. 
While pushing known class representations away from inactive prototypes facilitates knowledge preservation, it also impairs the discrimination of unknown classes.
As shown in \cref{Figure2b}, representations of unknown classes tend to be distributed within the narrow gaps between known prototypes rather than the broader open space where inactive prototypes reside.
As a result, the separability between known and unknown classes is severely compromised.
Besides, more adaptation is required when evolving unknowns to knowns, leading to excessive representation drift of previously learned classes.

To solve this challenge, we propose retentive angular representation learning (RARL) for IOSR. 
As illustrated in \cref{Figure2c}, inactive prototypes are not incorporated into optimization.
Instead, unknown representations are encouraged to be distributed in the broader open space opposite known prototypes (see \cref{Figure2d}).
This enhances the separability between known and unknown classes while avoiding excessive representation drift during knowledge updates.
Then, a virtual-intrinsic interaction (VII) training strategy is adopted to compact known representations while enforcing clear inter-class margins to improve the discriminability of decision boundaries.
Moreover, the impact of the multi-faceted data imbalance problem is rectified by imposing elastic constraints.
Finally, a new IOSR benchmark is established by comprehensively evaluating state-of-the-art OSR and CIL methods on the CIFAR100 and TinyImageNet datasets, where the proposed method consistently achieves leading performance.

The main contributions can be summarized as follows: 
\begin{itemize}
\item We propose retentive angular representation learning for IOSR, enabling stable expansion of recognizable scopes with the identified unknown classes.

\item We propose a virtual-intrinsic interactive training strategy. Without additional representation drift, both intra-class compactness and inter-class separability of intrinsic known representations are effectively improved.

\item We propose a stratified rectification strategy, which alleviates the impact of data imbalance between old/new and positive/negative class samples to refine OSR decision boundaries during knowledge updates.
\end{itemize}
\section{Related Work}
\label{sec:Related Work}

\textbf{Open set recognition} methods can be broadly classified into two categories:
(1) \textit{Discriminative} methods focus on optimizing the latent representations of known classes. 
Bendale \etal~\cite{bendale2016towards} brought discriminative methods from the classical machine learning era~\cite{scheirer2012toward,scheirer2014probability,cevikalp2016best} into the deep learning era by replacing the SoftMax layer with OpenMax.
Yang \etal~\cite{yang2018robust} introduced prototype learning to facilitate the discrimination between known and unknown classes, which has inspired subsequent extensions from the perspective of refining prototype geometry~\cite{miller2021class, liu2022orientational} and exploring adaptive mining mechanisms~\cite{lu2022pmal, tan2023prototype}.
(2) \textit{Generative} methods, on the other hand, estimate the distribution of unknown classes to improve their separation from known ones. Ge \etal~\cite{ge2017generative} extended OpenMax by employing generative adversarial networks (GANs) to synthesize data for novel classes, inspired hybrid architectures combining generation and discrimination~\cite{perera2020generative, kong2021opengan, chen2021adversarial, jang2023teacher,fu2025reason}. Alternative strategies employ latent reconstruction objectives \cite{yoshihashi2019classification} and conditional feature synthesis \cite{sun2020conditional}. Beyond this, Zhou \etal~\cite{zhou2021learning} introduced placeholders for both data and classifiers to fine-tune decision boundaries. %Despite these advancements, most methods are based on static assumptions. In dynamic environments, learning new classes can shift the representation, leading to forgetting of previous classes and greater confusion with unknown ones.

\textbf{Class incremental learning} addresses catastrophic forgetting through three principal paradigms:
(1) \textit{Data replay-based methods} mitigates forgetting by storing and revisiting prior samples. Current research focuses on optimal exemplar selection~\cite{rebuffi2017icarl, chaudhry2018riemannian, bang2021rainbow, fu2024class} and memory-efficient storage~\cite{zhou2022model, zhao2021memory}. Beyond direct instance storage, generative models can approximate class distributions and generate synthetic samples~\cite{jiang2021ib, sun2023exemplar}.  
(2) \textit{Parameter regularization-based methods} constrain key model parameters to prevent catastrophic forgetting. Following the seminal elastic weight consolidation (EWC) framework~\cite{kirkpatrick2017overcoming}, various importance estimation strategies have been proposed ~\cite{zenke2017continual, yang2019adaptive}. Li \etal~\cite{li2017learning} leveraged knowledge distillation to implicitly constrain parameters, inspiring a series of methods, including logic distillation ~\cite{smith2021always, hu2021distilling} and feature distillation ~\cite{hou2019learning,douillard2020podnet}. 
(3) \textit{Model rectification-based methods} bridge the gap between incremental models and static upper bounds through architectural adaptation. The dominant methods focus on classifier rectification~\cite{hou2019learning, zhao2020maintaining} and embedding space adjustment~\cite{zhou2022model, jie2022alleviating}. The neural collapse (NC) phenomenon has motivated some works to leverage the vertices of an ETF as classifier weights, directly optimizing the feature space towards an optimal structure~\cite{pernici2021class, he2024dyson}. Meanwhile, Wu \etal~\cite{wu2019large} proposed an alternative approach, indirectly rectifying the model by aligning its outputs.   

\section{Preliminaries}
\label{sec:Preliminaries}

\subsection{Problem Definition}
\label{sec:Problem definition}

In IOSR, the model is trained on sequences of tasks $\{D_{\text{train}}^1, \dots, D_{\text{train}}^T\}$ of length $T$. For each task $t\in\{1,\dots,T\}$, its training dataset $D_{\text{train}}^t = \{(x_i^t, y_i^t)\}_{i=1}^{n_t}$ contains $n_t$ instances $x_i^t$ with labels $y_i^t \in Y_{\text{train}}^t$. Crucially, the label spaces across tasks are mutually exclusive: $Y_{\text{train}}^{t_1} \cap Y_{\text{train}}^{t_2} = \emptyset$ for any $t_1 \neq t_2$. Only a small exemplar set $\mathcal{M}$, which stores samples from each old class of the previous $t-1$ tasks, is available during the $t$-th training task.
During evaluation, the test set $D_{\text{test}}^t$ for task $t$ comprises two distinct subsets: $D_K^t$ for known classes with
label space $Y_K^t=\bigcup_{j=1}^t Y_{\text {train }}^j$, and $D_U^t$ for unknown classes, the label space of which aligns with that of training data at the next task step, i.e., $Y_{U}^t = Y_{\text{train}}^{t+1}$.

The learning objective is to train a model $f: \mathcal{X} \rightarrow Y_{K}^t \cup Y_{U}^t$ by minimizing the following risk:
\begin{equation}
\underset{f \in \mathcal{H}}{\arg \min }\{ \left(\mathcal{R}_\epsilon\left(f, D_K^t\right)+\alpha \cdot \mathcal{R}_o\left(f, \mathcal{D}_U^t\right)\right\},
\end{equation}
where $\mathcal{H}$ denotes the hypothesis space, $\mathcal{R}_\epsilon$ and $\mathcal{R}_o$ represent the empirical risk and the open space risk, respectively.

\section{Method}
\label{sec:Method}

\begin{figure*}[t]
\centering
\includegraphics[width=\textwidth]{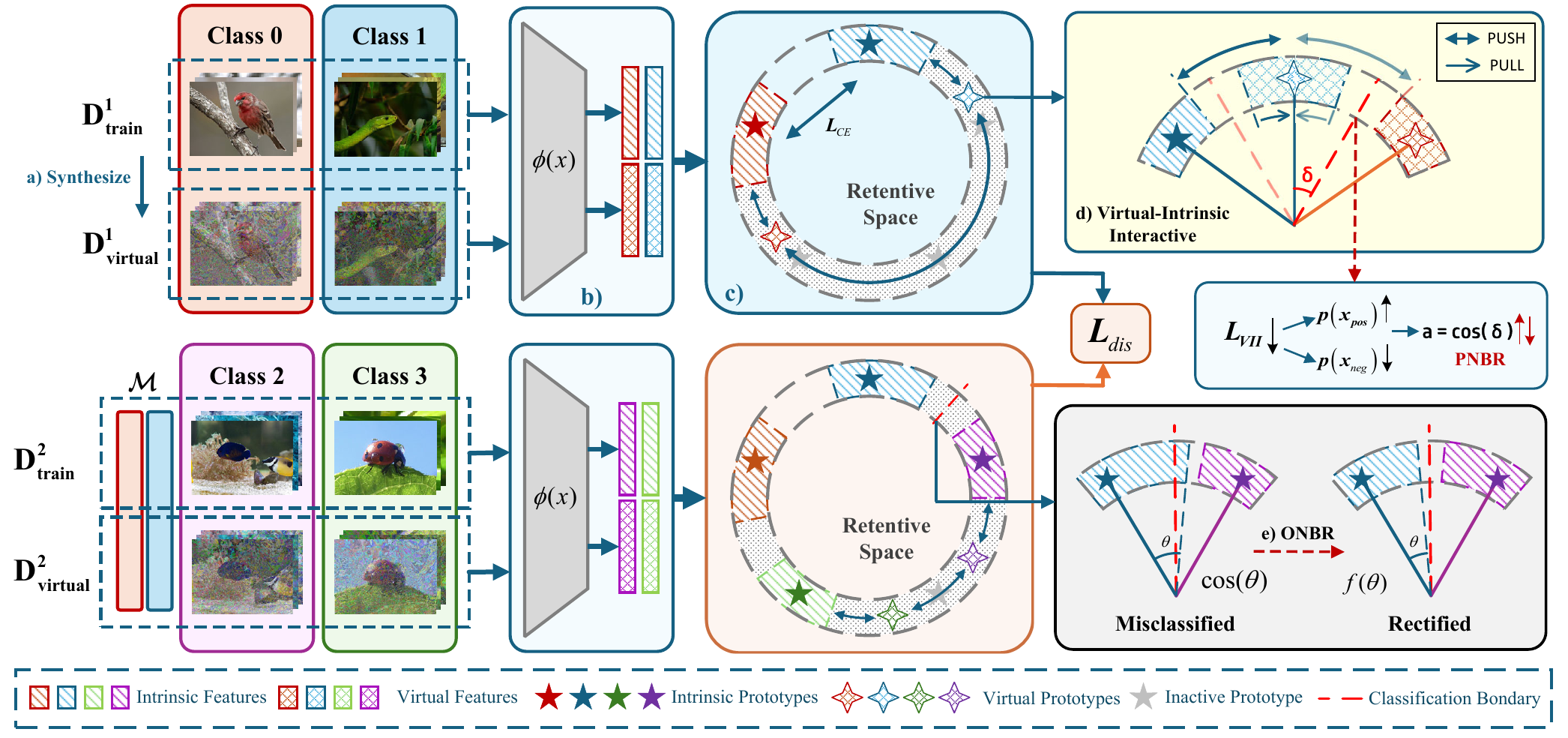}
\caption{An overall illustration of our method, which includes: (a) a virtual class synthesis module, (b) a feature extraction network, (c) a prototype-fixed retentive angular representation space, (d) a VII training scheme with PNBR, and (e) an ONBR module.
}
\label{Figure3}
\end{figure*}

\subsection{Framework Overview}

The overall framework of the RARL method is illustrated in \cref{Figure3}. We first pre-construct the retentive angular space by fixing the optimal geometric structure of the prototypes in the feature space.  

For the initial incremental task, we synthesize a virtual set $\mathbf{D}_{\text{virtual}}^1$ using the input training set $\mathbf{D}_{\text{train}}^1$ and extract features into the angular space via the backbone network $\phi(x)$. During classification, we preserve space for future classes while jointly optimizing intrinsic and virtual representations via the VII training strategy with PNBR, effectively clarifying the inter-class margins. For subsequent tasks, we leverage the retentive angular space as an optimization framework to guide new class features into the reserved regions, minimizing representation distortion. To address catastrophic forgetting caused by data imbalance, we introduce ONBR to rectify the boundary between old and new categories, further stabilizing the representations. At the end of each task, we save a small number of instances to the exemplar set $\mathcal{M}$ for rehearsal.

\subsection{Retentive Angular Space Construction} \label{fixed}

The NC phenomenon, discovered by Papyan \etal~\cite{papyan2020prevalence}, suggests that after training, final-layer features collapse into class-specific vertices of an ETF, maximizing angular separation. Pernici \etal~\cite{pernici2021class} leverage this by predefining class prototypes as an ETF to prevent prototype drift in incremental learning. 

Following this principle, the $K$ class prototypes, $m_i \in \mathbb{R}^d, i = 1, \ldots, K$, conform to the following equation:
\begin{equation}
\mathbf{M} = \sqrt{\frac{K}{K-1}} \mathbf{U} \left( \mathbf{I}_K - \frac{1}{K} \mathbf{1}_K \mathbf{1}_K^T \right),
\end{equation}
where $\mathbf{M} = \left[ m_1, \ldots, m_K \right]$, $\mathbf{I}_K$ is the identity matrix, $\mathbf{U} \in \mathbb{R}^{d \times K}$ is a randomly initialized rotation matrix satisfying $\mathbf{U}^T \mathbf{U} = \mathbf{I}_K$, and $\mathbf{1}_K$ is a $K$-dimensional all-ones vector. Since there is no prior knowledge about the total number of classes, we fix $d$ prototypes in advance based on the dimension of the last layer of the feature extraction network. Since the angles between all prototypes are equal, we can assign classes in any order and activate the corresponding prototypes when their classes appear.%For the $t$-th task, we refer to prototypes that have not yet been assigned a class as inactive prototypes.

% Based on the fixed prototypes, we use cosine similarity as a metric of distance. For a training instance $x_i^t$ in task $t$, the output prediction probability for the $c$-class is:
% \begin{equation}
% p_c^t\left(x_i^t\right) = \frac{\exp \left(\eta \left\langle \bar{m}_c, \bar{z_i}^t \right\rangle \right)}{\sum_{k=1}^K \exp \left( \eta \left\langle \bar{m}_k, \bar{z_i}^t \right\rangle \right)},
% \end{equation}
Based on the fixed prototypes, we use cosine similarity as a metric of distance. For a training instance $x_i^t$ in task $t$ with one-hot encoded label $\mathbf{y}_i^o = [y_{i1}^o, \dots, y_{iK}^o]$ (where $y_{ic}^o = 1$ if $x_i^t$ belongs to class $c$, and $0$ otherwise), the predicted probability for class $c$ and the cross-entropy loss are computed as follows:
\begin{align}
p_c^t\left(x_i^t\right) &= \frac{\exp \left(\eta \left\langle \bar{m}_c, \bar{z_i}^t \right\rangle \right)}{\sum_{k=1}^K \exp \left( \eta \left\langle \bar{m}_k, \bar{z_i}^t \right\rangle \right)}, \\
L_{ce}(x_i^t) &= -\sum_{c=1}^K y_{ic}^o \log \left( p_c^t(x_i^t) \right),
\end{align}

where $K$ is the number of classes for the current task, $\bar{v} = \frac{v}{\|v\|_2}$ denotes the $l_2$ normalized vector, and $\left\langle \bar{v}_a, \bar{v}_b \right\rangle = \bar{v}_a^T \bar{v}_b$ represents the cosine similarity between the pair of vectors. Following ~\cite{hou2019learning}, the learnable parameter $\eta$ is included to optimize the peakness of the SoftMax distribution. The denominator of the $p_c^t$ indicates that our cross-entropy loss doesn't involve inactive prototypes, enables us to construct a retentive angular space, ensuring better adaptability for unknown classes.

\subsection{Virtual-Intrinsic Interactive Training Scheme} \label{VII}

To further enhance the discriminability of decision boundaries, we draw inspiration from ~\cite{kalantidis2020hard} to synthesize virtual instances and facilitate interactions with intrinsic classes. The synthesis process for virtual classes is illustrated in \cref{negcat_syn}:
\begin{equation}
x_i^V = \lambda x_i + \frac{(1 - \lambda)}{K_{batch} - 1} \sum_{j \neq i} x_j,
\label{negcat_syn}
\end{equation}
where $\lambda=0.5$, and $K_{batch}$ represents the number of intrinsic classes in a batch. Since each virtual instance is synthesized primarily based on an intrinsic instance, virtual classes remain distinct from one another while share similarities with corresponding intrinsic classes. To utilize this semantic relation, and without prior knowledge of the virtual classes' specific positions in feature space, we assign new, autonomously learnable prototypes (instead of fixed intrinsic prototypes) to each virtual class for classification, thereby ensuring semantic stability during training. For a virtual instance $x_i^V$, the classification loss is computed as \cref{Lneg}:
\begin{equation}
L_V(x_i^V) = - \sum_{c=1}^K y_{ic}^V \log \left(p_c^V(x_i^V)\right),
\label{Lneg}
\end{equation}
where $y_{ic}^{V}$ represents the one-hot label of the virtual class $c$, and $p_c^{V}$ denotes the predicted probability, which is obtained by computing the cosine similarity with the virtual prototype. Inclusion of this classification loss can maintain the divisibility for virtual features and prevent them from losing their characteristic semantic patterns and degrading into a single noise cluster in open space, thereby preserving their connection to their corresponding intrinsic classes.

Since we preserve the semantic patterns of the virtual class, its feature distribution remains close to that of its corresponding intrinsic class. To further enhance the discrimination between intrinsic classes, we introduce VII loss to assist training, reinforcing the separation between ambiguous virtual classes and intrinsic class features. For a virtual class $c$, the cosine similarity between the feature of instance $x_i^V$ and the prototype $m_c^V$ of class $c$ is first transformed into a probability using the sigmoid function:
\begin{equation}
p_c^V\left(x_i^V\right) = \frac{1}{1 + \exp \left(- \eta \left\langle \bar{m}_c^V, \bar{\phi}\left(x_i^V\right) \right\rangle \right)}
\end{equation}
The total VII loss is formulated as follows:
\begin{equation}
\mathcal{L}_{VII}(\mathcal{B}) = \frac{1}{|\mathcal{B}|} \sum_{\mathbf{x}_i \in \mathcal{B}} \ell_{VII}(\mathbf{x}_i),
\end{equation}
where $\mathcal{B} = \mathcal{B}^t \cup \mathcal{B}^V$ is a batch containing intrinsic instances $\mathcal{B}^t$ and virtual instances $\mathcal{B}^V$, and the per-instance loss is:
\begin{equation}
\scalebox{0.8}{$
\ell_{VII}(x_i) =
\begin{cases}
-\log p_y^V(x_i) - \sum\limits_{c=1}^{K} \log\big(1 - p_c^V(x_i)\big), & x_i \in \mathcal{B}^V \\[0.5em]
-\log\big(1 - p_y^V(\mathbf{x}_i)\big), & x_i \in \mathcal{B}^t
\end{cases}
$}
\label{Loss_VII_revised}
\end{equation}
with $y$ denoting the true class of $\mathbf{x}_i$.
The VII loss $L_{VII}(x)$ induces distinct gradient dynamics for three critical probability terms during optimization:

\begin{itemize}
    \item \textbf{Virtual instances ($\mathbf{x}_i \in \mathcal{B}^V$)}:
    \begin{align*}
        \frac{\partial \ell_{VII}}{\partial p_y^V} &= -\frac{1}{p_y^V(\mathbf{x}_i)} \\
        \frac{\partial \ell_{VII}}{\partial p_c^V} &= \frac{1}{1 - p_c^V(\mathbf{x}_i)} \quad \forall c \in \{1,...,K_V\}
    \end{align*}
    \item \textbf{Intrinsic instances ($\mathbf{x}_i \in \mathcal{B}^t$)}:
    \begin{align*}
        \frac{\partial \ell_{VII}}{\partial p_y^V} &= \frac{1}{1 - p_y^V(\mathbf{x}_i)}
    \end{align*}
\end{itemize}
This creates a triple separation dynamic:
\begin{align*}
a) \small \underbrace{p_y^V(\mathbf{x}_i^V) \nearrow 1}_{\text{virtual alignment}} \quad
b)\small\underbrace{p_y^V(\mathbf{x}_y^t) \searrow 0}_{\text{intrinsic-virtual separation}} \quad 
c)\small\underbrace{p_c^V(\mathbf{x}_i^V) \searrow 0}_{\text{virtual discrimination}}
\end{align*}

$a)$ Maximizes the similarity probability between virtual features and their corresponding prototype, thereby promoting intra-class feature compactness.

$b)$ Ensures intrinsic representations are repelled from the low confidence region where the corresponding virtual prototype is located. 

$c)$ Enforces mutual exclusion between distinct virtual class prototypes, further preserving the semantic patterns of virtual classes.

\subsection{Stratified Rectification Strategy}  \label{SRS}

While $L_{VII}$ facilitates the separation between virtual and intrinsic features, the pull-to-push ratio is $1 : (K - 1) + 1$ (one virtual class against other virtual classes and it's intrinsic class), leading to an imbalance where the push effect is stronger than pull. Consequently, the feature distribution of virtual classes shifts farther from all prototypes, thereby distorting the intended feature structure.

Furthermore, each incremental task exhibits a quantitative imbalance between rehearsal data and new training data, resulting in significant classification boundary drift.

To address these two problems, a stratified rectification strategy is proposed. It consists of two components: a) positive and negative boundary rectification (PNBR) and b) old and new boundary rectification (ONBR).

\begin{figure}[]
    \centering
    \begin{subfigure}{0.49\columnwidth} % 指定子图的宽度
        \centering
        \includegraphics[width=\linewidth]{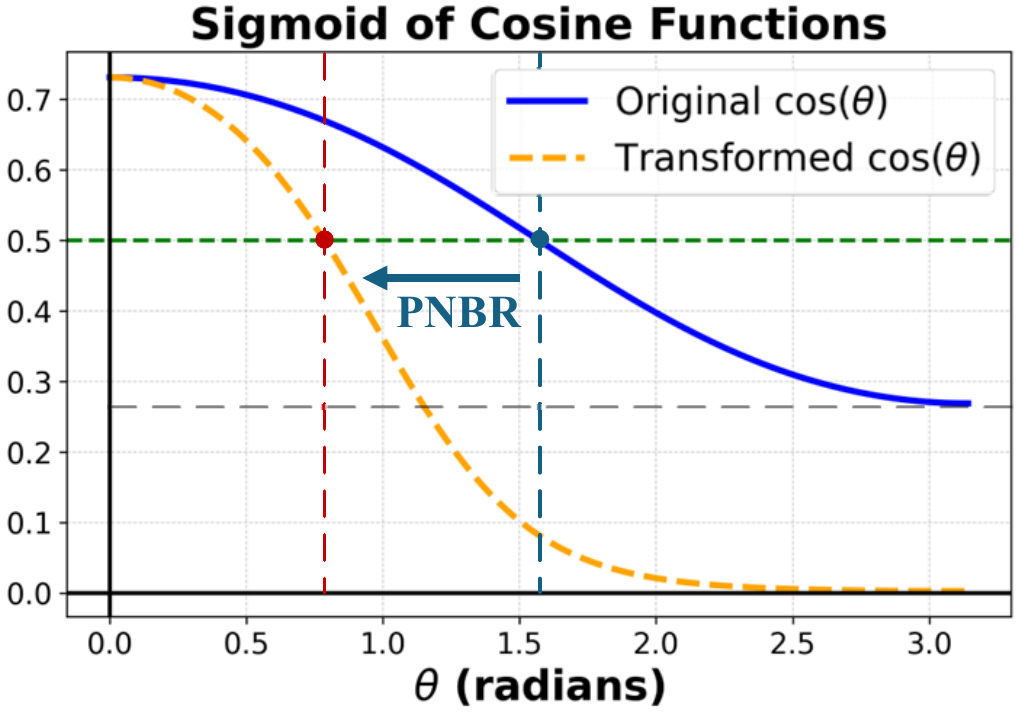}
        \caption{}
        \label{Figure4a}
    \end{subfigure}
    \begin{subfigure}{0.49\columnwidth}
        \centering
        \includegraphics[width=\linewidth]{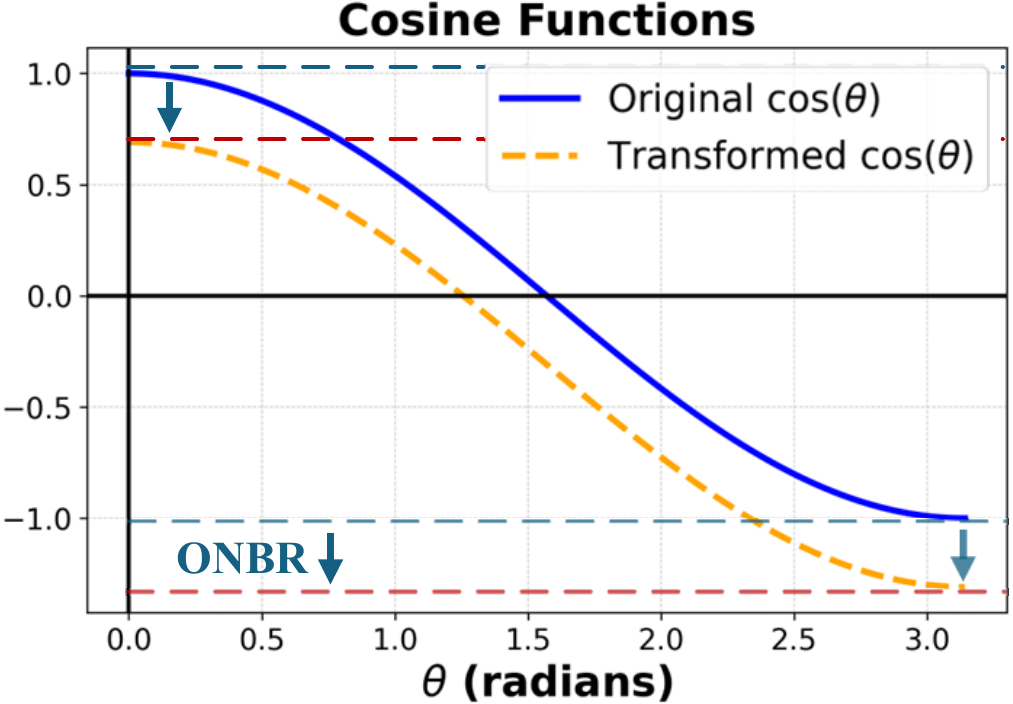}
        \caption{}
        \label{Figure4b}
    \end{subfigure}
    \caption{(a) Sigmoid function variation after using PNBR. (b) Cosine function variation after using ONBR.}
    \label{Figure4}
\end{figure}

\textbf{a) PNBR}. To alleviate the effects of the imbalance between positive (one virtual class) and negative instances (other virtual classes and intrinsic class) in $L_{VII}$, we transform the cosine similarity to adaptively adjust the classification boundary as \cref{PNBR}:
\begin{equation}
f(\theta) = \frac{1}{1-a} (\cos(\theta) - a),
\label{PNBR}
\end{equation}
where $\theta$ denotes the angle between the two vectors and $a \in (0,1)$ is a learnable parameter. The original cosine similarity $\cos(\theta)=\left\langle \bar{m}_c^V, \bar{\phi}\left(x_i\right) \right\rangle$ in $L_{VII}$ is transformed into $\frac{1}{1-a} (\left\langle \bar{m}_c^V, \bar{\phi}\left(x_i\right) \right\rangle - a)$. 

The transformed cosine similarity sigmoid curve, along with the original sigmoid curve, is depicted in \cref{Figure4a}. The transformation exhibits a strictly subtractive effect on raw cosine similarities for all $a > 0$. During optimization, this will cause two effects:

\begin{enumerate}
    
   \item \textbf{Positive-Class Focus} (virtual instances):
    The gradient dynamics for virtual-class prototypes reveal a dual amplification mechanism:
    {\setlength{\abovedisplayskip}{3pt}
     \setlength{\belowdisplayskip}{3pt}
     \begin{flalign*}
        \frac{\partial \mathcal{L}}{\partial \cos(\theta_{pos})} &= (p_{pos}^V - 1) \cdot \frac{1}{1-a}, \\
        |\text{gradient}| &\propto \underbrace{(1 - p_{pos}^V)}_{\substack{\text{Increases as}\\ f(\theta_{pos}) \searrow}} \cdot \underbrace{\frac{1}{1-a}}_{\substack{\text{Increases as}\\ a \nearrow}} \nearrow
     \end{flalign*}
    }
    
    Crucially, the reduction in $p_{pos}^V$ caused by $f(\theta_{pos})$ creates a feedback loop:
    \begin{itemize}
        \item Lower $f(\theta_{pos}) \Rightarrow$ Lower $p_{pos}^V \Rightarrow$ Stronger gradient
        \item The scaling factor $\frac{1}{1-a}$ further magnifies this effect
        \item Results in accelerated convergence of virtual features to their prototypes
    \end{itemize}

    \item \textbf{Negative-Class Relaxation} (intrinsic/other virtual):
    For non-target classes, the gradient exhibits opposing behavior:
    {\setlength{\abovedisplayskip}{3pt}
     \setlength{\belowdisplayskip}{3pt}
    \begin{align*}
    \frac{\partial \mathcal{L}}{\partial \cos(\theta_{neg})} &= p_{neg}^V\cdot\frac{1}{1-a}, \\
    |\text{gradient}| &\propto \underbrace{p_{neg}^V}_{\substack{\text{Decreases as}\\ f(\theta_{neg}) \searrow}} \cdot \underbrace{\frac{1}{1-a}}_{\substack{\text{Increases as}\\ a \nearrow}}\searrow
    \end{align*}
    }
    The reduction in $p_{neg}^V$ caused by $f(\theta_{neg})$ creates a feedback loop:
    \begin{itemize}
        \item Lower $f(\theta_{neg}) \Rightarrow$ Lower $p_{neg}^V \Rightarrow$ Weaker gradient
        \item While $\frac{1}{1-a}$ increases with $a$, its effect is outweighed by $p_{neg}^V$ reduction
        \item Creates "soft repression" rather than hard separation from negative classes
    \end{itemize}
\end{enumerate}
Eventually we will adaptively learn a suitable $a$, rectifying the boundary between the positive and negative classes by slowing down the negative class constraints with additional constraints on the positive class.

\textbf{b) ONBR.} To mitigate the catastrophic forgetting problem caused by the imbalance between old and new data, we introduce a rectification named ONBR for the cosine similarity of old classes as shown in \cref{Figure4b}. Unlike PNBR, ONBR introduces a hyperparameter $A$ and uses $A \cdot \frac{\pi}{2}$ instead of $a$. During optimization, the cosine similarity for old classes is reduced, causing features with the same angle as their corresponding prototypes to incur a larger loss. This encourages a stronger focus on constraining old class features, thereby mitigates the excessive drift of classification boundaries towards new classes.
\subsection{Overall classification loss}
Following \cite{hou2019learning}, we also incorporate the less forget constraint to minimize the variation in feature distribution during the incremental learning process. The corresponding loss function is defined as:
\begin{equation}
L_{dis} = 1 - \left\langle \bar{Z}_{t-1}, \bar{Z_t} \right\rangle,
\end{equation}
where $\bar{Z}_{t-1}$ represents the normalized features of the old model. Our overall loss function can then be summarized as:
\begin{equation} 
L_{Total} = L_{ce} + L_V + \lambda_{VII}L_{VII} + \lambda_{dis} L_{dis},
\end{equation}
where $\lambda_{VII}$ is the weight for the VII loss (set to 0.01), and $\lambda_{dis}$ is the weight for the less-forget loss.

\section{Experiment}\label{sec:Experiment}
\begin{table*}[t]
\caption{Experimental results on the CIFAR100 dataset}
\label{CIFAR}
\centering
\setlength{\tabcolsep}{1.7mm}
\renewcommand\arraystretch{0.5}
\begin{tabular}{lccccccccccccc}
\toprule 
\midrule
\multirow{3}{*}{Method} & \multirow{3}{*}{Scenario} & \multicolumn{6}{c}{Base 20 with 8 Steps} & \multicolumn{6}{c}{Base 20 with 4 Steps}\\ 
\cmidrule(lr){3-8} \cmidrule(lr){9-14}
& & \multicolumn{2}{c}{ACC} &  \multicolumn{2}{c}{AUROC} & \multicolumn{2}{c}{OSCR} & \multicolumn{2}{c}{ACC} & \multicolumn{2}{c}{AUROC} &  \multicolumn{2}{c}{OSCR} \\
\cmidrule(lr){3-4} \cmidrule(lr){5-6} \cmidrule(lr){7-8} \cmidrule(lr){9-10}
\cmidrule(lr){11-12} \cmidrule(lr){13-14} 
& & Avg&Last &Avg&Last&Avg&Last&Avg&Last&Avg&Last &Avg&Last\\
\midrule
SoftMax & / & 47.32& 36.2& 61.17& 59.71& 36.22& 27.45& 57.00& 42.41& 65.76& 58.67& 45.37& 31.93 \\
ODL~\cite{liu2022orientational} & OSR & 45.61& 35.54& 59.08& 59.38& 33.08& 25.65& 56.06& 38.98& 63.79& 56.21& 43.08& 28.09 \\ 
GCPL~\cite{yang2018robust} & OSR & 48.26& 35.44& 55.09& 52.63& 30.76& 20.59& 58.69& 44.11& 61.11&52.79& 41.9& 27.29 \\ 
ARPL~\cite{chen2021adversarial} & OSR & 46.79& 35.93& 63.12& 58.72& 36.73& 27.69& 57.37& 41.09& 68.01& 61.55& 45.69& 31.51 \\
CAC~\cite{miller2021class} & OSR & 48.12& 37.51& 59.61& 57.70& 34.96& 26.65& 58.44& 42.44& 64.65&56.99& 44.82& 30.21 \\ 
MEDAF~\cite{wang2024exploring} & OSR & 36.81& 24.19& 59.36& 59.23& 28.07& 18.65& 48.37& 32.54& 63.51&55.00& 37.84& 23.71 \\
iCaRL~\cite{rebuffi2017icarl} & CIL & 49.41& 37.62& 60.61& 57.78& 37.12& 27.64& 66.42&54.55& 68.48& 62.90& 53.17& 42.62 \\
WA~\cite{zhao2020maintaining} & CIL & 52.12& 42.53& 63.44& 61.22& 41.12& 33.72& 60.48& 48.00& 67.59& 62.74& 48.79& 38.43 \\ 
DR~\cite{yang2023neural} & CIL & 49.17& 38.02& 59.36& 58.10& 35.64& 27.93& 59.56& 44.56& 63.98&58.37& 45.16 & 32.05 \\ 
CWD~\cite{shi2022mimicking} & CIL & 56.85& 45.1& 63.57& 59.57& 43.76& 34.27& 65.82& 52.10& 67.59&61.71& 51.75& 40.21 \\
LUCIR~\cite{hou2019learning} & CIL & 57.18& 46.17& 64.15& 58.02& 44.55& 34.85& 66.55& 53.86& 68.51&60.73& 52.83& 40.67 \\
% DeepIncrement  & OpenIncrement  & 53.5& 36.4& 62.3& 58.1& 34.3& 21.3& 60.7& 42.2& 65.4& 57.9& 40.8& 24.3 \\
\midrule \rowcolor{gray!10}
\textbf{RARL} & \textbf{IOSR} & \textbf{62.56}& \textbf{48.53}& \textbf{66.84}& \textbf{61.69}& \textbf{48.86}& \textbf{37.37}& \textbf{70.63}& \textbf{57.71}& \textbf{68.77}& \textbf{64.93}& \textbf{54.91}&  \textbf{44.72} \\  
\midrule
\bottomrule
\end{tabular}
\end{table*}

\begin{table*}[t]
\caption{Experimental results on the TinyImageNet dataset}
\label{TinyImageNet}
\centering
\setlength{\tabcolsep}{1.7mm}
\renewcommand\arraystretch{0.5}
\begin{tabular}{lccccccccccccc}
\toprule 
\midrule
\multirow{3}{*}{Method} & \multirow{3}{*}{Scenario} & \multicolumn{6}{c}{Base 50 with 10 Steps} & \multicolumn{6}{c}{Base 60 with 7 Steps}\\ 
\cmidrule(lr){3-8} \cmidrule(lr){9-14}
& & \multicolumn{2}{c}{ACC} &  \multicolumn{2}{c}{AUROC} & \multicolumn{2}{c}{OSCR} & \multicolumn{2}{c}{ACC} & \multicolumn{2}{c}{AUROC} &  \multicolumn{2}{c}{OSCR} \\
\cmidrule(lr){3-4} \cmidrule(lr){5-6} \cmidrule(lr){7-8} \cmidrule(lr){9-10}
\cmidrule(lr){11-12} \cmidrule(lr){13-14} 
& & Avg&Last &Avg&Last&Avg&Last&Avg&Last&Avg&Last &Avg&Last\\
\midrule
SoftMax & / & 33.31& 22.62& 58.66& 55.44& 25.73& 16.32& 35.85& 25.19& 60.08& 56.99& 28.28 & 18.88
\\
ODL~\cite{liu2022orientational} & OSR& 28.31& 17.32& 56.03& 49.91& 21.76& 11.70& 31.07 & 18.78 & 57.88 & 53.68 & 24.66 & 13.75  \\ 
GCPL~\cite{yang2018robust} & OSR & 30.87& 17.08& 53.58& 48.06& 21.58& 10.20 & 34.89 & 21.90 & 55.94 & 49.24 & 25.49 & 13.59 \\ 
ARPL~\cite{chen2021adversarial} & OSR & 32.11& 22.63& 59.97& 60.43& 25.51& 18.02 & 34.31 & 23.4 & 61.41 & 59.55 & 28.02 & 18.96 \\
CAC~\cite{miller2021class} & OSR & 29.38& 11.79& 57.16& 52.82& 22.81& 9.67 & 32.05 & 19.99 & 58.17 & 55.54 & 25.61 & 15.19\\ 
MEDAF~\cite{wang2024exploring} & OSR & 34.32 & 24.58 & 60.60 & 57.30 & 26.88 & 18.29 & 36.93 & 26.77 & 61.24 & 57.99 & 29.00 & 19.94 \\
iCaRL~\cite{rebuffi2017icarl} & CIL & 44.82&33.51& 62.83& 63.48& 36.06& 27.38 & 47.40 & 36.74 & 64.05 & 62.34 & 38.62 & 29.77\\
WA~\cite{zhao2020maintaining} & CIL & 41.62& 29.65& 62.62& 58.92& 34.23& 24.21 & 43.47 & 31.03 & 64.14 & 61.51 & 36.15 & 25.53\\ 
DR~\cite{yang2023neural} & CIL & 33.84& 23.51& 58.15& 54.26& 25.85& 16.67 & 36.71 & 25.1 & 59.37 & 57.27 & 28.61 & 19.07
 \\ 
CWD~\cite{shi2022mimicking} & CIL & 44.05& 31.69& 63.14& 64.13& 35.61& 26.44 & 47.14 & 34.51 & 63.73 & 63.11 & 38.18 & 28.23\\
LUCIR~\cite{hou2019learning} & CIL & 47.56& 36.41& 64.79& 65.36& 39.08& 30.24 & 50.96 & 39.04 & 66.39 & \textbf{66.25} & 42.33 & 32.65 \\
% DeepIncrement  & OpenIncrement  & 53.5& 36.4& 62.3& 58.1& 34.3& 21.3& 60.7& 42.2& 65.4& 57.9& 40.8& 24.3 \\
\midrule\rowcolor{gray!10}
\textbf{RARL} & \textbf{IOSR} & \textbf{52.93}& \textbf{39.94}& \textbf{65.46}& \textbf{66.29}& \textbf{42.69}& \textbf{33.13}& \textbf{55.49}& \textbf{42.26}& \textbf{67.27}& 65.64& \textbf{45.08}& \textbf{34.48} \\
\midrule
\bottomrule
\end{tabular}
\end{table*}

\begin{figure*}[t]
\centering
\includegraphics[width=0.9\textwidth]{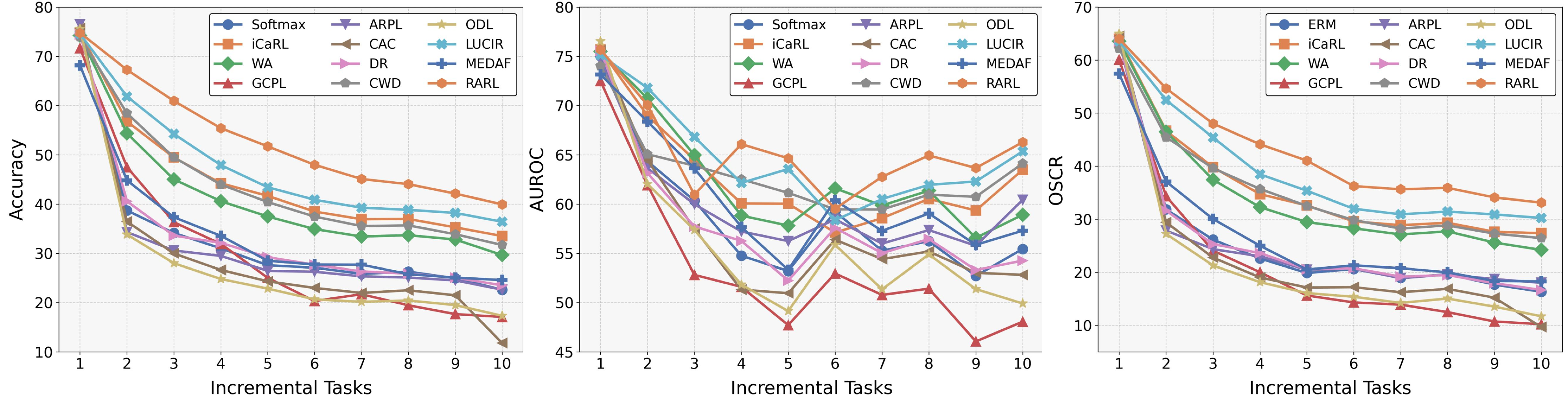}
\caption{Line plot of per-task performance on TinyImageNet under Base 50 with 10 Steps setting.
}
\label{Tiny_result}
\end{figure*}
\subsection{Implementation Details}

\subsubsection{Datasets} 
We evaluate on two benchmark datasets:
(1) CIFAR100 (100 classes): Initial task contains 20 randomly selected classes; the remaining classes are equally split into either 7 or 3 incremental tasks.
(2) TinyImageNet (200 classes): Initial task contains either 50 or 60 randomly selected classes, with the remaining classes equally split into 9 or 7 incremental tasks, respectively.

\subsubsection{Training details} 
Our experiments are conducted on NVIDIA RTX 3090 GPUs using the PyTorch framework. We employ ResNet-34 as the backbone, optimize with the SGD. The initial learning rate is 0.1, with a momentum of 0.9 and a weight decay of $5 \times 10^{-4}$. The model is trained for 160 epochs, with the learning rate reduced by a factor of 0.1 at epochs 80 and 120. Consistent with most CIL methods, we randomly select and store 20 instances per class.

% \begin{table}[]
% \caption{Experimental results on the TinyImageNet dataset}
% \label{TinyImageNet}
% \setlength{\tabcolsep}{1.7mm}
% \renewcommand\arraystretch{1}
% \begin{tabular}{lcccccc}
% \toprule
% \multirow{3}{*}{Method} &\multicolumn{6}{c}{Base 50 with 10 Steps} \\ \cmidrule(lr){2-7} 
% & \multicolumn{2}{c}{ACC} &  \multicolumn{2}{c}{AUROC} & \multicolumn{2}{c}{OSCR} \\
% \cmidrule(lr){2-3} \cmidrule(lr){4-5} \cmidrule(lr){6-7} 
% &Avg&Last &Avg&Last&Avg&Last\\
% \midrule
% SoftMax & 33.31& 22.62& 58.66& 55.44& 25.73& 16.32
% \\
% iCaRL & 44.82&33.51& 62.83& 63.48& 36.06& 27.38
% \\
% WA & 41.62& 29.65& 62.62& 58.92& 34.23& 24.21 
% \\ 
% ODL & 28.31& 17.32& 56.03& 49.91& 21.76& 11.70
% \\ 
% GCPL & 30.87& 17.08& 53.58& 48.06& 21.58& 10.20
% \\ 
% ARPL & 32.11& 22.63& 59.97& 60.43& 25.51& 18.02
% \\ 
% CAC & 29.38& 11.79& 57.16& 52.82& 22.81& 9.67
% \\ 
% DR & 33.84& 23.51& 58.15& 54.26& 25.85& 16.67
% \\ 
% CWD & 44.05& 31.69& 63.14& 64.13& 35.61& 26.44
% \\ 
% Lucir & 47.56& 36.41& 64.79& 65.36& 39.08& 30.24
% \\ 
% \midrule
% \textbf{Ours} & \textbf{52.93}& \textbf{39.94}& \textbf{65.46}& \textbf{66.29}& \textbf{42.69}& \textbf{33.13}
% \\   
% \bottomrule
% \end{tabular}
% \end{table}

\subsubsection{Comparative methods} 
We reproduce several representative CIL and OSR methods within our experimental framework, including: iCaRL~\cite{rebuffi2017icarl}, WA~\cite{zhao2020maintaining}, ODL~\cite{liu2022orientational}, LUCIR~\cite{hou2019learning}, CWD~\cite{shi2022mimicking}, GCPL~\cite{yang2018robust}, ARPL~\cite{chen2021adversarial}, DR~\cite{yang2023neural}, CAC~\cite{miller2021class} and MEDAF~\cite{wang2024exploring}. 
We adopt cross-entropy training with randomly sampled rehearsal instances as our baseline approach. 
%For CIL methods, we follow their standard training protocols (with data replay). During testing, we evaluate both unknown class discrimination and closed-set performance. For OSR methods, we train them on the stored exemplar set and new class data at each incremental step, evaluated with the same metrics as CIL methods for fair comparison.

\subsubsection{Evaluation protocol} 
We evaluate both closed-set and open set performance under dynamic settings using classification accuracy, AUROC~\cite{neal2018open}, and OSCR~\cite{dhamija2018reducing} as metrics. Performance is reported for both the final task and the average across all tasks.

\begin{table*}[]
\caption{Results of ablation experiments on the CIFAR100 dataset}
\label{table_ablation}
\centering
\setlength{\tabcolsep}{1.4mm}
\renewcommand\arraystretch{0.5}
\begin{tabular}{lcccccccccccccc}
\toprule\midrule 
\multirow{2}{*}{Methods} &  \multirow{2}{*}{$L_{d i s}$} & \multirow{2}{*}{$L_V$}  & \multirow{2}{*}{$L_{VII}$} & \multirow{2}{*}{PNBR} & \multirow{2}{*}{ONBR} &\multicolumn{8}{c}{Tasks} &\multirow{2}{*}{Average}
\\ \cmidrule(lr){7-14}
& & & & & & 1 & 2 & 3 & 4 & 5 & 6 & 7 & 8\\ 
\midrule
SoftMax (ALL) &   &   &   &   &   & 68.71 & 37.51& 35.23 & 32.96& 29.48 & 31.86 & 27.16 & 27.89 & 36.35 
\\
w/ $L_{d i s}$ & \checkmark &   &  &   &   & 67.77 & 52.47 & 49.34 & 44.46 & 41.65 & 37.37 & 34.29 & 33.79 & 45.14 
\\
w/ $L_V$ & \checkmark & \checkmark  &  &   &   & 70.07 &  52.97& 48.83 & 44.48& 41.91 & 38.63 & 35.69 & 34.68 & 45.91 
\\
w/o PNBR & \checkmark & \checkmark & \checkmark &   &   & 69.10 & 51.79 & 49.02 &  44.15& 42.35 & 39.08 & 35.03
 & 35.21 & 45.72 
\\
w/ PNBR & \checkmark & \checkmark & \checkmark & \checkmark  &   & \textbf{71.65} &  51.80 & 48.23 & 45.08& 42.44 & 38.88 & 35.63 & 34.96 & 46.08 
\\
w/ ONBR & \checkmark & &  &   & \checkmark  & 67.77 & 51.23& 52.48 &  46.69& 46.66 & 40.69 & 38.52 & 35.99 & 47.51 
\\
\midrule \rowcolor{gray!10}
\textbf{RARL} &  \checkmark & \checkmark & \checkmark & \checkmark  & \checkmark  & \textbf{71.65}&  \textbf{53.09}& \textbf{52.53} &  \textbf{48.65} & \textbf{48.27} & \textbf{41.31} & \textbf{39.95} & \textbf{37.15} & \textbf{49.08}
\\
\midrule
\bottomrule
\end{tabular}
\end{table*}

\subsection{Performance Analysis}

The results of all methods across different settings are presented in \cref{CIFAR} and \cref{TinyImageNet}. It can be observed that the CIL approach outperforms the OSR approach in both open set and closed set metrics across all experimental setups. This phenomenon is more intuitive in \cref{Tiny_result}. Which can be attributed to the fact that the primary objective of CIL methods is to mitigate catastrophic forgetting of old classes in closed-set scenarios within dynamic environments. According to the theory in~\cite{vaze2021open}, the discriminative ability of a model for closed sets is directly proportional to its ability for open sets. Since existing OSR methods are designed for static scenarios, the severe forgetting of old classes in incremental settings leads to a gradual degradation of the model's discriminative ability.

\begin{figure}[]
    \centering
    \begin{subfigure}{0.9\linewidth} % 指定子图的宽度
        \centering
        \includegraphics[width=\linewidth]{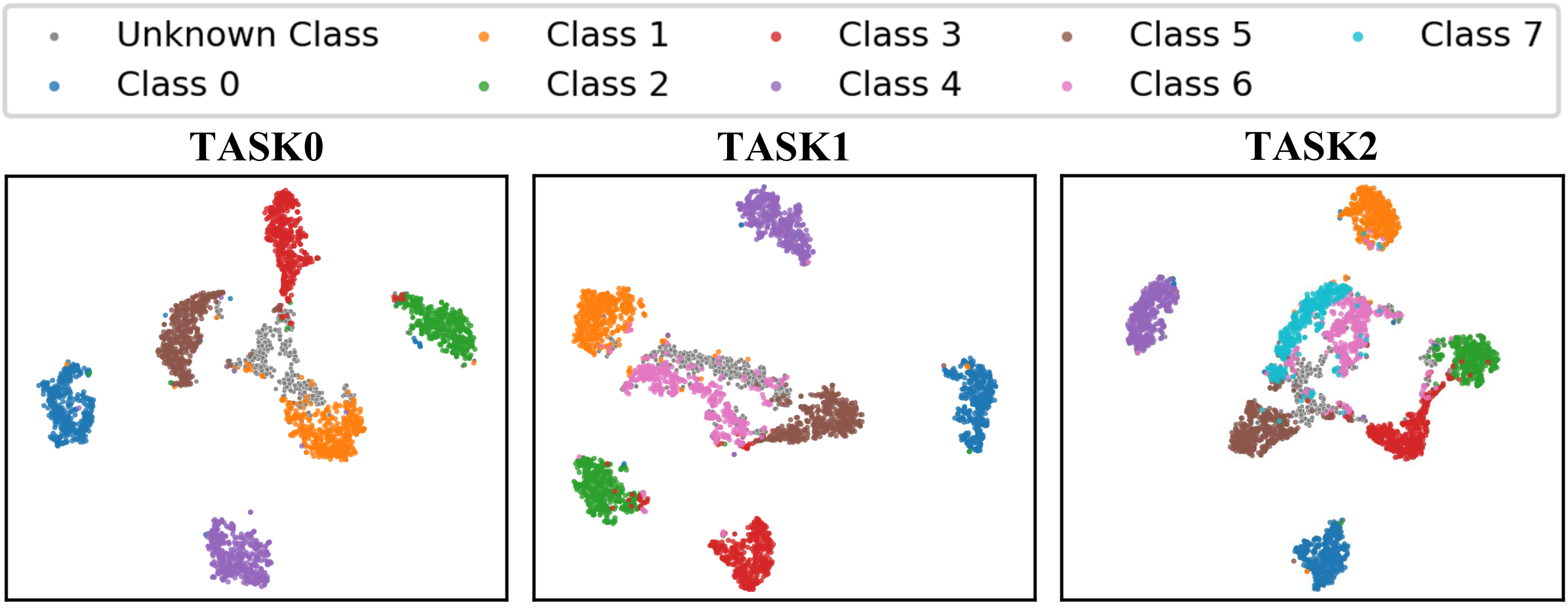}
        \caption{LUCIR}
        \label{Figure7a}
    \end{subfigure}
    \\
    \begin{subfigure}{0.9\linewidth}
        \centering
        \includegraphics[width=\linewidth]{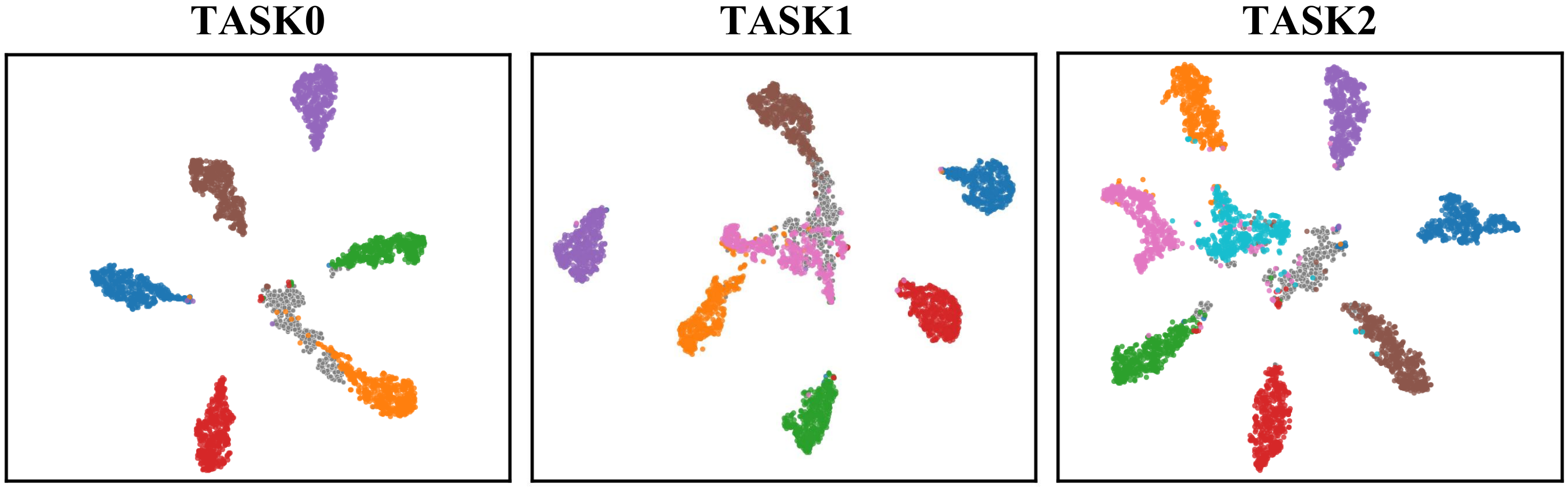}
        \caption{RARL}
        \label{Figure7b}
    \end{subfigure}
    \caption{t-SNE visualization of learned features on CIFAR-100 under a simple incremental setting with partial classes.}
    \label{Figure7}
\end{figure}

On both CIFAR100 and TinyImageNet, LUCIR demonstrates suboptimal performance. Compared to LUCIR, RARL improves the average accuracy by $4.08\% \sim 5.38\%$, the average AUROC by $0.26\% \sim 2.69\%$, and the average OSCR by $1.74\% \sim 4.31\%$. \cref{Figure7} visualizes the feature distributions learned by RARL and LUCIR using t-SNE. It can be observed that LUCIR maintains a certain degree of feature stability, indicating that the cosine similarity used by both LUCIR and RARL supports more consistent representations compared to euclidean distance-based metrics during incremental learning. However, our method yields a more distinct and discriminative feature structure, suggesting that RARL enhances class separability and better distinguishes unknown classes in dynamic scenarios, thus offering a targeted solution for IOSR.

\subsection{Ablation Study}

\textbf{Influence of Different Components.} The proposed RARL has four main components: virtual class classification loss ($L_V$), virtual-intrinsic interaction loss ($L_{VII}$), PNBR and ONBR. We conduct ablation experiments on the CIFAR-100 dataset using the Base 20 with 8 Steps setup and evaluate performance using the OSCR metric, as shown in \cref{table_ablation}. SoftMax (ALL) represents the loss where all fixed prototypes are considered. 

When all fixed prototypes are included, the open set capacity is lowest in the first task, confirming that including future prototypes impairs open set performance. In constructing our retentive angular space, adding only $L_V$ improves OSCR by $1.36\%$, showing that virtual instances alone enhance open set ability. However, applying $L_{\text{VII}}$ without PNBR reduces performance, as the imbalance between positive and negative classes overcompacts features, leaving insufficient space for unknown classes. With PNBR, the first task achieves optimal performance, increasing OSCR by $0.36\%$ and validating PNBR's effectiveness. Adding only ONBR leads to greater improvements in dynamic settings, showing that it effectively mitigates forgetting due to class imbalance. The highest performance is achieved when all components are applied together. This aligns with our goal: each component plays a crucial role, and together they enhance open set capability while maintaining performance in dynamic environments.

\section{Conclusion}

In this paper, we tackle the IOSR problem by proposing RARL. Specifically, after releasing space around inactive prototypes, we adopt a VII training strategy to refine inter-class margins. Moreover, to mitigate the adverse effects of data imbalance, we introduce the stratified rectification strategy, which adjusts decision boundaries both within VII loss (for positive and negative classes) and during incremental learning (for new and old classes). Additionally, we propose a new experimental benchmark that better simulates real-world dynamic environments. In this benchmark, unknown classes from the current incremental task later become training classes in subsequent tasks, creating a continually evolving mix of known and unknown classes. Extensive experiments demonstrate that our method outperforms all baselines, achieving superior results across both closed set and open set metrics. These results prove that RARL provides a robust solution to the IOSR problem.
{\small

\bibliographystyle{ieee_fullname}
}

\end{document}